\documentclass{article}

\PassOptionsToPackage{numbers, compress}{natbib}

 \usepackage[dblblindworkshop, final]{neurips_2025}
\workshoptitle{Multimodal Representation Learning for Healthcare}

\usepackage[utf8]{inputenc} 
\usepackage[T1]{fontenc}    
\usepackage{hyperref}       
\usepackage{url}            
\usepackage{booktabs}       
\usepackage{amsfonts}       
\usepackage{nicefrac}       
\usepackage{microtype}      
\usepackage{xcolor}         
\usepackage{amsmath}
\usepackage{graphicx}
\usepackage{subcaption}  

\title{Towards Multimodal Representation Learning in Paediatric Kidney Disease}

\author{%
  Ana Durica \\
  Institute of Health Informatics \\
  University College London \\
  London, United Kingdom \\
  \texttt{ana.durica.21@ucl.ac.uk} \\
  \And
  John Booth\\
  Data Research, Innovation and Virtual Environments Unit\\
  Great Ormond Street Hospital\\
  London, United Kingdom \\
  \texttt{john.booth@gosh.nhs.uk} \\
  \And
  Ivana Drobnjak \\
  Department of Computer Science \\
  University College London \\
  London, United Kingdom \\
  \texttt{i.drobnjak@ucl.ac.uk} \\
}

\begin{document}

\maketitle

\begin{abstract}
Paediatric kidney disease varies widely in its presentation and progression, which calls for continuous monitoring of renal function. Using electronic health records collected between 2019 and 2025 at Great Ormond Street Hospital, a leading UK paediatric hospital, we explored a temporal modelling approach that integrates longitudinal laboratory sequences with demographic information. A recurrent neural model trained on these data was used to predict whether a child would record an abnormal serum creatinine value within the following thirty days. Framed as a pilot study, this work provides an initial demonstration that simple temporal representations can capture useful patterns in routine paediatric data and lays the groundwork for future multimodal extensions using additional clinical signals and more detailed renal outcomes.
\end{abstract}

\section{Introduction}

Paediatric kidney disease varies widely in its clinical presentation and rate of progression \cite{Cirillo2023CKDchildren}, which requires continuous monitoring of renal function to detect early deterioration. Serum creatinine, the standard marker of kidney function, reflects the glomerular filtration rate and underpins diagnosis and staging \cite{kdigo2024ckd, schwartz2009egfr}. Anticipating short-term changes in creatinine could support several paediatric nephrology decisions, including early identification of kidney injury, assessment of treatment response, and monitoring of disease progression.

Great Ormond Street Hospital (GOSH) is a leading paediatric centre with an advanced electronic health record (EHR) system and a trusted digital research environment \cite{GOSH_DRIVE_ImpactReport_2023}. The dataset used in this study contains multi-year clinical and biochemical information, including demographics, laboratory results, hospital episodes and procedures, all linked at the patient level. Representation learning provides a way to combine such heterogeneous inputs into patient embeddings that capture temporal and cross-modal structure. Recurrent neural networks (RNNs) have shown promising performance in adult EHR tasks, including chronic kidney disease progression and heart-failure prediction \cite{lei2018_rnndae, zhu2023_ckd_rnn}, but similar work in paediatric nephrology is limited.

In this study, we develop a simple pipeline to organise EHR data into a format suitable for RNNs. As an initial prediction task, we focus on whether a patient will record an abnormal creatinine value within the following 30 days, using only two modalities: routine laboratory tests and basic demographic information. This work serves as an initial step toward multimodal representation learning in paediatric kidney disease, with the longer-term aim of incorporating additional data sources and more clinically meaningful renal outcomes.

\section{Methodology}

This study developed a pipeline for combining longitudinal laboratory data and demographic features to predict whether a patient would record an abnormal creatinine value within the following 30 days. The workflow included: (1) cohort selection and outcome definition, (2) temporal structuring of longitudinal records, and (3) sequence modelling with an RNN and evaluation of the learned representations.

\subsection{Cohort selection and outcome definition}

Data were extracted from the EHR for all patients with at least one nephrology episode between 2019 and 2025 ($n = 1{,}032$), of whom $38$ (3.5\%) died during follow-up. Follow-up was defined from the first to the last recorded serum creatinine, or to the date of death if applicable. To form a 30-day prediction window ending at the last follow-up time ($t_{\text{end}}$), patients were included if they (1) had at least three creatinine measurements on separate days before the window and (2) if deceased, had at least one measurement within the final 30 days of life. These criteria yielded a final cohort of $n = 826$, including $17$ patients (2.1\%) who died but had a valid measurement within the window. The prediction target was the occurrence of any abnormal serum creatinine value within the 30-day window. Abnormality status was taken directly from EHR laboratory flags without additional processing. Patients with no abnormal values were labelled as $0$ ($n = 370$, 44.8\%), and those with one or more abnormal values as $1$ ($n = 456$, 55.2\%).

\subsection{Data extraction and preprocessing}

For the selected cohort, a longitudinal dataset was created by identifying all valid creatinine measurements prior to the 30-day prediction window and using their dates as event time points. At each time point, all laboratory tests performed on the same calendar day were extracted, and their presence and abnormality status were recorded. Fifteen clinically relevant laboratory markers were included, each represented by two binary indicators (presence and abnormality), giving 30 features, together with age and sex as static variables. Absent tests were encoded as \texttt{presence = 0} and \texttt{abnormal = 0}, while performed tests had \texttt{presence = 1} with the abnormality flag reflecting test status. This produced temporally ordered sequences of laboratory information for each patient, forming the input to the time-series model.

\subsection{Sequence modelling}

Given a sequence of patient records \( X = (x_1, x_2, \dots, x_n) \), where \( x_t \) is the multi-hot laboratory feature vector at time \( t \), the model aimed to learn a latent representation \( c \). Sequences were limited to 100 time points, with shorter sequences left-padded with zeros to preserve temporal order. A gated recurrent unit (GRU) encoder \cite{Chung2014GRU} processed \( X \) to capture temporal dependencies, and the resulting embedding \( c \) was concatenated with demographic features and passed to a linear classifier to predict 30-day outcomes. The model was trained using a stratified 70/10/20 train–validation–test split to maintain outcome balance.

\subsection{Evaluation}

Model performance was evaluated on the held-out 20\% test set. Discrimination was assessed using the area under the ROC curve (AUC), with 95\% confidence intervals estimated from 2{,}000 bootstrap resamples. Using a 0.5 decision threshold, we computed and displayed the confusion matrix to illustrate classification behaviour. Predicted probabilities were obtained from sigmoid-transformed logits, and test-set embeddings were visualised using t-SNE.

\section{Results}

We present the cohort’s temporal and laboratory structure, the model’s performance in predicting abnormal creatinine within the following 30 days, and an exploratory visualisation of the learned embeddings.

\subsection{Cohort and temporal characteristics}

\begin{figure}[h]
  \centering
  \begin{subfigure}[t]{0.49\linewidth}
    \includegraphics[width=\linewidth]{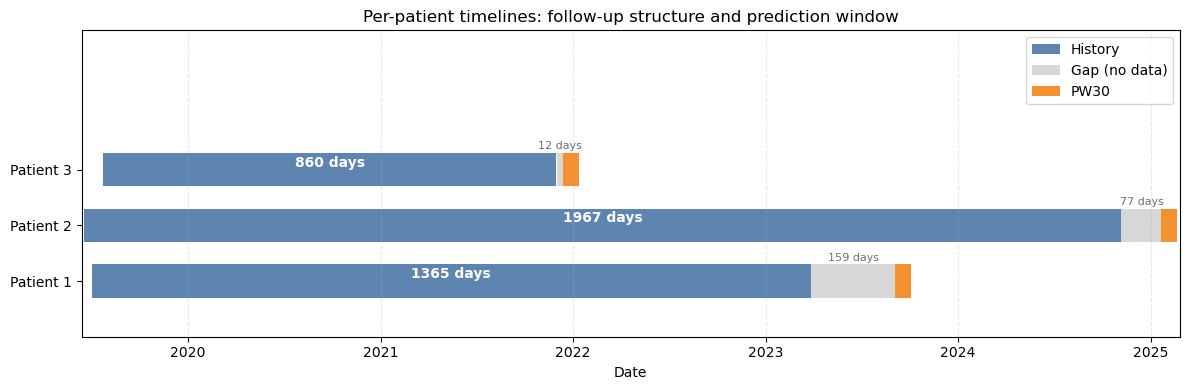}
    \subcaption{Follow-up timelines for example patients. Blue bars show the available pre-window history, grey indicates periods without recorded laboratory data, and orange marks the fixed 30-day prediction window. Timelines are shown in calendar time, corresponding to each patient’s recorded laboratory events.}
    \label{fig:temporal_structure_a}
  \end{subfigure}
  \hfill
  \begin{subfigure}[t]{0.49\linewidth}
    \includegraphics[width=\linewidth]{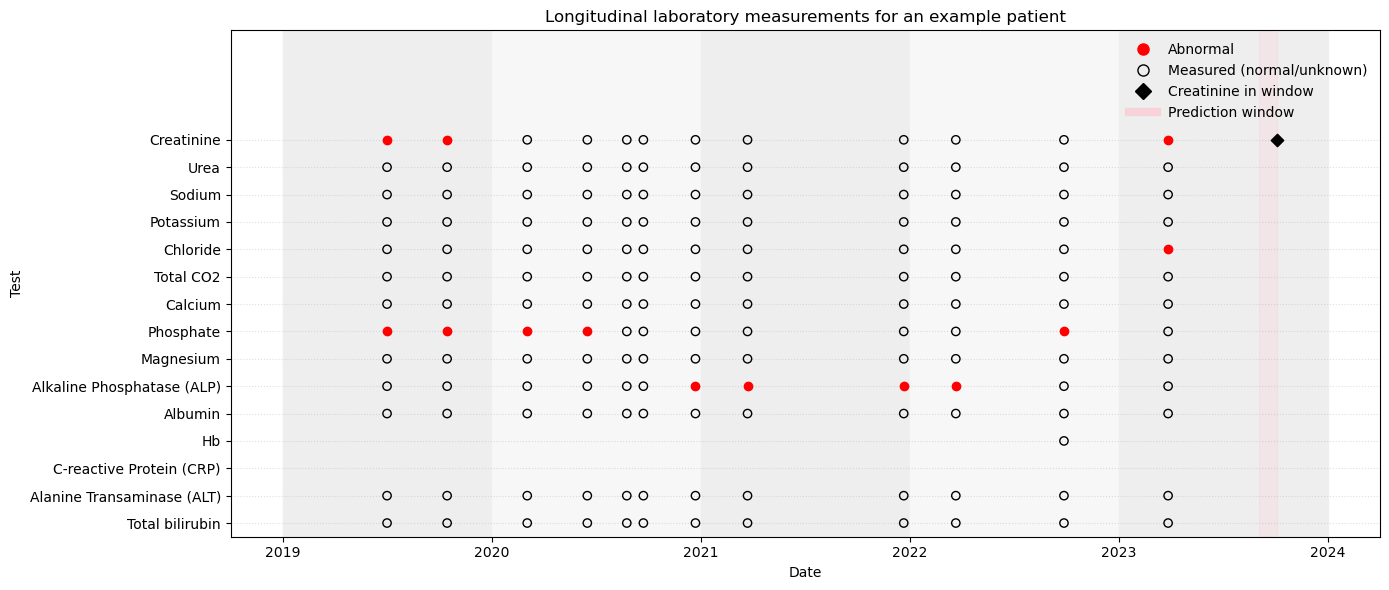}
    \subcaption{Example patient’s laboratory measurements over time. Red circles indicate abnormal results, open circles denote measured tests with non-abnormal values, and missing markers indicate that a test was not performed at that event time point. Alternating grey bands show calendar years, and the pink region marks the 30-day prediction window.}
    \label{fig:temporal_structure_b}  
  \end{subfigure}
  \caption{Overview of longitudinal data used for temporal modelling.}
  \label{fig:temporal_structure}
\end{figure}

Figure~\ref{fig:temporal_structure} illustrates the cohort’s longitudinal patterns: follow-up durations varied substantially across patients, and laboratory measurements were recorded at irregular intervals with occasional data gaps. Consequently, the 30-day prediction window frequently followed periods without recent laboratory observations, mirroring real clinical practice.

\subsection{Predictive performance}

\begin{figure}[h]
  \centering
  \begin{subfigure}[t]{0.48\linewidth}
    \centering
    \includegraphics[width=\linewidth]{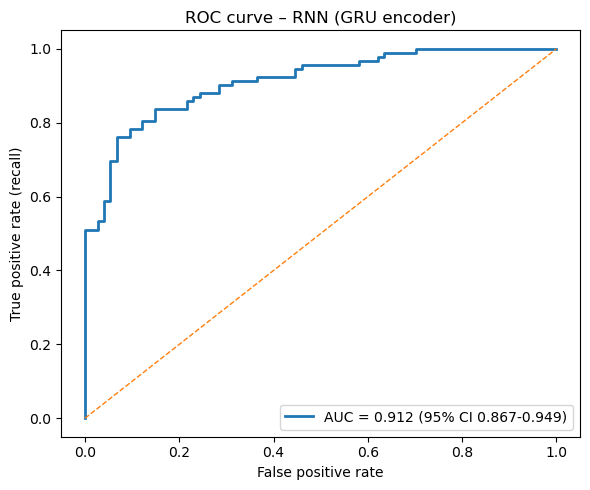}
    \subcaption{ROC curve for the GRU model on the test set.}
    \label{fig:roc_curve}
  \end{subfigure}
  \hfill
  \begin{subfigure}[t]{0.48\linewidth}
    \centering
    \includegraphics[width=\linewidth]{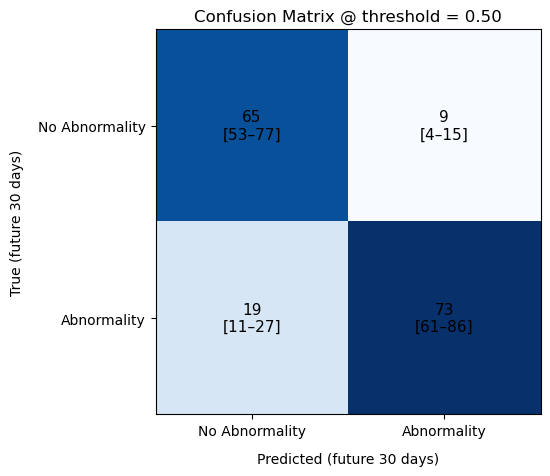}
    \subcaption{Confusion matrix at a 0.5 decision threshold.}
    \label{fig:confusion_matrix}
  \end{subfigure}
  \caption{Predictive performance of the GRU model. Bootstrapped 95\% confidence intervals are shown for both the ROC curve and the confusion-matrix cell counts.}
  \label{fig:predictive_performance}
\end{figure}

The GRU-based model was trained to predict abnormal creatinine within the following 30-day window. Figure~\ref{fig:predictive_performance} shows that the model achieved good discrimination, providing preliminary support for this short-term prediction task.

\subsection{Embedding visualisation}

\begin{figure}[h]
  \centering
  \includegraphics[width=0.55\linewidth]{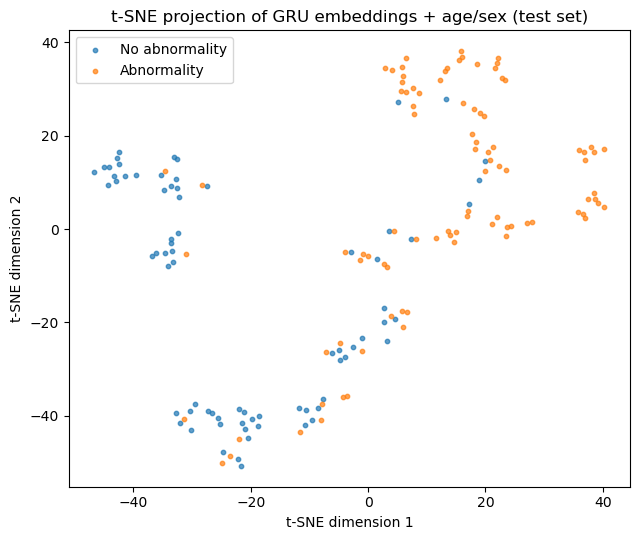}
  \caption{t-SNE projection of GRU embeddings for the test set. Each point represents a patient embedding coloured by 30-day outcome.}
  \label{fig:embeddings_tsne}
\end{figure}

The GRU embeddings from the test set were projected into two dimensions using t-SNE (Figure~\ref{fig:embeddings_tsne}). 

This exploratory visualisation shows partially separated regions for patients with and without future abnormal creatinine values, suggesting that the model captured preliminary predictive signal.

\section{Discussion}

This work provides a preliminary demonstration that temporal modelling of longitudinal EHR data can support the near-term prediction of creatinine abnormalities in paediatric patients. Using a GRU encoder applied to laboratory event sequences, together with static demographic features, we achieved good discrimination in predicting whether a patient would have an abnormal creatinine test within 30 days. These findings suggest that recurrent models can extract clinically relevant temporal signals from laboratory histories.

Several simplifications shaped this pilot. Sequences were treated as step-indexed events without explicit timing, despite irregular sampling. Laboratory features were anchored to creatinine measurement dates, blending physiological information with patterns of clinical monitoring. Tests were encoded as binary indicators of presence and abnormality, without quantitative values. These design choices enabled a tractable first implementation but reduce the clinical detail captured in the representations.

Although the current model uses only two input modalities (laboratory data and basic demographics), the underlying GOSH cohort contains a far broader set of signals. Multi-year trajectories include laboratory, medication, diagnosis, vital-sign and procedural data, each with distinct temporal structure. Future work will extend this framework to multimodal and time-aware models, and to more clinically meaningful renal endpoints such as sustained creatinine elevation, acute kidney injury, treatment response or post-transplant outcomes. These directions aim to make fuller use of the depth of paediatric EHR data and support more proactive renal care.

\begin{ack}
This research was supported by UKRI (EP/S021612/1) through the UCL Centre for Doctoral Training in AI-enabled Healthcare Systems and by GOSH Children’s Charity Award VS0618. Research at GOSH NHS Foundation Trust and the UCL Great Ormond Street Institute of Child Health is supported by the NIHR GOSH Biomedical Research Centre. The views expressed are those of the authors and not necessarily those of the NHS, the NIHR or the Department of Health.
\end{ack}

\bibliographystyle{plainnat}
\bibliography{refs}

\end{document}